\documentclass[10pt,twocolumn,letterpaper]{article}

\usepackage{cvpr}              
\definecolor{cvprblue}{rgb}{0.21,0.49,0.74}
\usepackage[pagebackref,breaklinks,colorlinks,allcolors=cvprblue]{hyperref}

\def\paperID{18183} 
\def\confName{CVPR}
\def\confYear{2025}

\title{Network Inversion and Its Applications}

\author{Pirzada Suhail\\
IIT Bombay\\
India\\
{\tt\small psuhail@iitb.ac.in}
\and
Hao Tang\\
Peking University\\
China\\
{\tt\small haotang@pku.edu.cn}
\and
Amit Sethi\\
IIT Bombay\\
India\\
{\tt\small asethi@iitb.ac.in}
}

\begin{document}
\maketitle
\begin{abstract}
Neural networks have emerged as powerful tools across various applications, yet their decision-making process often remains opaque, leading to them being perceived as "black boxes."  This opacity raises concerns about their interpretability and reliability, especially in safety-critical scenarios. Network inversion techniques offer a solution by allowing us to peek inside these black boxes, revealing the features and patterns learned by the networks behind their decision-making processes and thereby provide valuable insights into how neural networks arrive at their conclusions, making them more interpretable and trustworthy. This paper presents a simple yet effective approach to network inversion using a meticulously conditioned generator that learns the data distribution in the input space of the trained neural network, enabling the reconstruction of inputs that would most likely lead to the desired outputs. To capture the diversity in the input space for a given output, instead of simply revealing the conditioning labels to the generator, we encode the conditioning label information into vectors and intermediate matrices and further minimize the cosine similarity between features of the generated images. Additionally, we incorporate feature orthogonality as a regularization term to boost image diversity which penalises the deviations of the Gram matrix of the features from the identity matrix, ensuring orthogonality and promoting distinct, non-redundant representations for each label.  The paper concludes by exploring immediate applications of the proposed network inversion approach in interpretability, out-of-distribution detection, and training data reconstruction.
\end{abstract}    
\section{Introduction}
\label{sec:intro}
Neural networks have become indispensable in a wide array of applications, ranging from image recognition and natural language processing to autonomous driving and medical diagnostics. Despite their remarkable performance, the decision-making processes within these networks often remain elusive, earning them the moniker "black boxes." This opacity poses significant challenges, particularly in scenarios where interpretability and reliability are paramount, such as in safety-critical applications.

The lack of transparency in neural network decision-making has raised concerns about their trustworthiness and the ability to diagnose and rectify errors. As these models become increasingly integrated into critical systems, there is a growing demand for techniques that can shed light on the internal workings of these networks. Network inversion offers a promising solution by enabling us to inspect and understand the features and patterns that neural networks learn during their training processes.

Network inversion techniques provide a mechanism to reveal the internal representations and decision-making pathways of neural networks. By inverting the network, we can reconstruct inputs that are likely to produce specific outputs, thereby gaining insights into the network’s learned data distribution and feature extraction processes. This capability is crucial for enhancing the interpretability and transparency of neural networks, making them more trustworthy and reliable.

In this paper, we present a network inversion technique that learns the input space corresponding to different classes within a classifier using a single conditioned generator trained to generate a diverse set of samples from the input space with desired labels guided by a combination of losses including cross-entropy, KL Divergence, cosine similarity and feature orthogonality. To ensure the generator learns a diverse set of inputs, we alter the conditioning from simple labels to vectors and matrices that encode the label information. This diversity is reinforced through the application of heavy dropout during the generation process and by minimizing the cosine similarity between the features of the generated images. Additionally, we incorporate feature orthogonality as a regularization term, by penalizing deviations of the Gram matrix of the features from the identity matrix. The orthogonality loss combined with cosine similarity helps achieve a more varied set of generated images, each corresponding to different conditioning vectors. 

Our methodology not only increases the diversity of the generated inputs but also provides deeper insights into the decision-making processes of neural networks. By revealing the hidden patterns and features that influence network predictions, we gain a more comprehensive understanding of neural network behavior. This understanding is crucial for several applications, including improving interpretability, enhancing safety, and boosting adversarial robustness.

We also demonstrate how our approach can be used to generate interpretable decision boundaries from the features of inverted images, providing a clearer view of the network's classification strategies. Furthermore, we explore the applications of inversion in reconstructing training data by exploiting unique properties of the training data relative to the classifier. We also employ inversion for out-of-distribution (OOD) detection by retraining the classifier with an additional class for “garbage” inverted samples, which aids in identifying data that do not belong to the training distribution.
\section{Related Works}
\label{sec:related}

The concept of neural network inversion has garnered significant attention as a method for visualizing and understanding the internal mechanisms of neural networks. Inversion seeks to identify input patterns that closely approximate a given output target, thereby revealing the information processing capabilities embedded within the network's weights. These methods reveal important insights into how models represent and manipulate data, offering a pathway to expose the latent structure of neural networks. Early research on inversion for multi-layer perceptrons in \citep{KINDERMANN1990277}, derived from the back-propagation algorithm, demonstrates the utility of this method in applications like digit recognition highlighting that while multi-layer perceptrons exhibit strong generalization capabilities—successfully classifying untrained digits—they often falter in rejecting counterexamples, such as random patterns. 

Subsequently \citep{784232} expanded on this idea by proposing evolutionary inversion procedures for feed-forward networks that stands out for its ability to identify multiple inversion points simultaneously, providing a more comprehensive view of the network’s input-output relationships. The paper \citep{SAAD200778} explores the lack of explanation capability in artificial neural networks (ANNs) and introduces an inversion-based method for rule extraction to calculate the input patterns that correspond to specific output targets, allowing for the generation of hyperplane-based rules that explain the neural network's decision-making process. \citep{Wong2017NeuralNI} addresses the problem of inverting deep networks to find inputs that minimize certain output criteria by reformulating network propagation as a constrained optimization problem and solving it using the alternating direction method of multipliers. Model Inversion attacks in adversarial settings are studied in \citep{10.1145/3319535.3354261}, where an attacker aims to infer training data from a model's predictions by training a secondary neural network to perform the inversion, using the adversary's background knowledge to construct an auxiliary dataset, without access to the original training data.

The paper \citep{NEURIPS2020_373e4c5d} presents a method for tackling data-driven optimization problems, where the goal is to find inputs that maximize an unknown score function by proposing Model Inversion Networks (MINs), which learn an inverse mapping from scores to inputs, allowing them to scale to high-dimensional input spaces. While \citep{ansari2022autoinverseuncertaintyawareinversion} introduces an automated method for inversion by focusing on the reliability of inverse solutions by seeking inverse solutions near reliable data points that are sampled from the forward process and used for training the surrogate model. By incorporating predictive uncertainty into the inversion process and minimizing it, this approach achieves higher accuracy and robustness. 

The traditional methods for network inversion often rely on gradient descent through a highly non-convex loss landscape, leading to slow and unstable optimization processes. To address these challenges, recent work by \citep{liu2022landscapelearningneuralnetwork} proposes learning a loss landscape where gradient descent becomes efficient, thus significantly improving the speed and stability of the inversion process. Similarly \cite{suhail2024network} proposes an alternate approach to inversion by encoding the network into a Conjunctive Normal Form (CNF) propositional formula and using SAT solvers and samplers to find satisfying assignments for the constrained CNF formula. While this method, unlike optimization-based approaches, is deterministic and ensures the generation of diverse input samples with desired labels. However, the downside of this approach lies in its computational complexity, which makes it less feasible for large-scale practical applications.

Our approach to neural network inversion aims to strike a balance between computational efficiency and the diversity of generated inputs by using a carefully conditioned generator trained to learn the data distribution in the input space of a trained neural network. The conditioning information is encoded into vectors in a concealed manner to enhance the diversity of the generated inputs by avoiding easy shortcut solutions. This method is further enhanced through the application of heavy dropout during the generation process and the minimization of cosine similarity between a batch of the features of the generated images. This combination of techniques ensures a diverse representation of the input space for any given output, thereby addressing the limitations of previous methods. Additionally, our approach is computationally less expensive compared to search-based SAT solvers, making it more feasible for practical applications.
\section{Methodology}
\label{sec:method}

Our approach to Network Inversion uses a single carefully conditioned generator that learns diverse data distributions in the input space of the trained classifier.
\subsection{Classifier}
In this paper inversion and reconstruction is performed on a classifier which includes convolution and fully connected layers as appropriate to the classification task. We use standard non-linearity layers like Leaky-ReLU \citep{xu2015empiricalevaluationrectifiedactivations} and Dropout layers \citep{JMLR:v15:srivastava14a} in the classifier for regularisation purposes to discourage memorisation. The classification network is trained on a particular dataset and then held in evaluation mode for the purpose of inversion.

\subsection{Generator}
The images in the input space of the classifier will be generated by an appropriately conditioned generator. The generator builds up from a latent vector by up-convolution operations to generate the image of the given size. While generators are conventionally conditioned on an embedding learnt of a label for generative modelling tasks, we given its simplicity, observe its ineffectiveness in network inversion and instead propose more intense conditioning mechanism using vectors and matrices. 

\subsubsection{Label Conditioning}
Label Conditioning of a generator is a simple approach to condition the generator on an embedding learnt off of the labels each representative of the separate classes. The conditioning labels are then used in the cross entropy loss function with the outputs of the classifier. While Label Conditioning can be used for inversion, the inverted samples do not seem to have the  diversity that is expected of the inversion process due to the simplicity and varying confidence behind the same label.

\subsubsection{Vector Conditioning}
In order to achieve more diversity in the generated images, the conditioning mechanism of the generator is altered by encoding the label information into an \(N\)-dimensional vector for an \(N\)-class classification task. The vectors for this purpose are randomly generated from a normal distribution and then soft-maxed to represent an input conditioning distribution for the generated images. The \(\text{argmax}\) index of the soft-maxed vectors now serves as the de facto conditioning label, which can be used in the cross-entropy loss function without being explicitly revealed to the generator.

\subsubsection{Intermediate Matrix Conditioning}
Vector Conditioning allows for a encoding the label information into the vectors using the argmax criteria. This can be further extended into Matrix Conditioning which apparently serves as a better prior in case of generating images and allows for more ways to encode the label information for a better capture of the diversity in the inversion process. In its simplest form we use a Hot Conditioning Matrix in which an \(NXN\) dimensional matrix is defined such that all the elements in a given row and column (same index) across the matrix are set to one while the rest all entries are zeroes. The index of the row or column set to 1 now serves as the label for the conditioning purposes. The conditioning matrix is concatenated with the latent vector intermediately after up-sampling it to \(NXN\) spatial dimensions, while the generation upto this point remains unconditioned.

\subsubsection{Vector-Matrix Conditioning}
Since the generation is initially unconditioned in Intermediate Matrix Conditioning, we combine both vector and matrix conditioning, in which vectors are used for early conditioning of the generator upto \(NXN\) spatial dimensions followed by concatenation of the conditioning matrix for subsequent generation. The argmax index of the vector, which is the same as the row or column index set to high in the matrix, now serves as the conditioning label.

\begin{figure*}[t]
\centering
\includegraphics[width=0.9\textwidth]{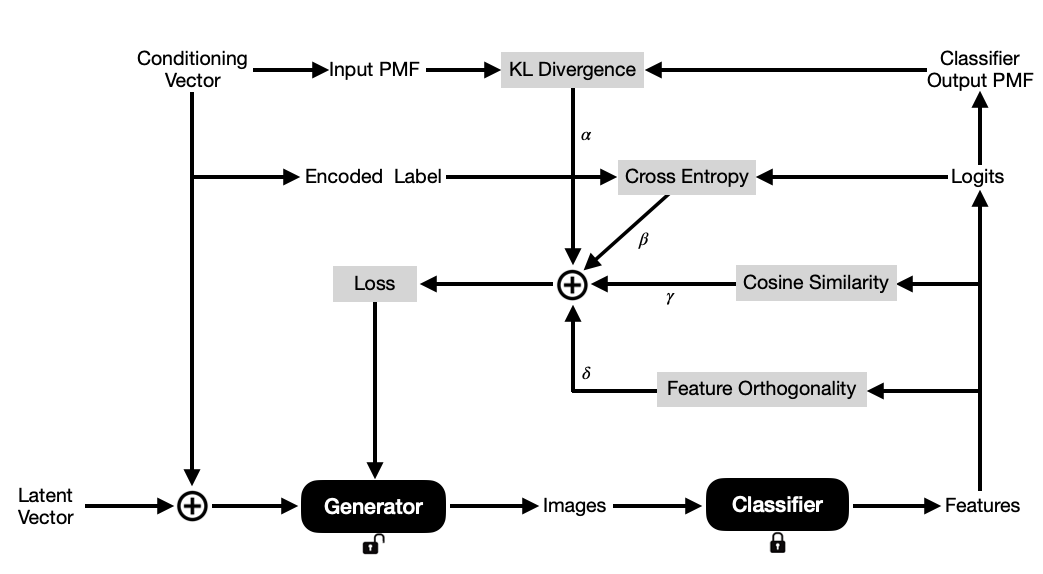} 
\caption{Proposed Approach to Network Inversion}
\label{1}
\end{figure*}

\subsection{Network Inversion}
The main objective of Network Inversion is to generate images that when passed through the classifier will elicit the same label as the generator was conditioned to. Achieving this objective through a straightforward cross-entropy loss between the conditioning label and the classifier’s output can lead to mode collapse, where the generator finds shortcuts that undermine diversity. With the classifier trained, the inversion is performed by training the generator to learn the data distribution for different classes in the input space of the classifier as shown schematically in Figure \ref{1} using a combined loss function \( \mathcal{L}_{\text{Inv}} \) defined as:
\[
\mathcal{L}_{\text{Inv}} = \alpha \cdot \mathcal{L}_{\text{KL}} + \beta \cdot \mathcal{L}_{\text{CE}} + \gamma \cdot \mathcal{L}_{\text{Cosine}} + \delta \cdot \mathcal{L}_{\text{Ortho}}
\]
where \( \mathcal{L}_{\text{KL}} \) is the KL Divergence loss, \( \mathcal{L}_{\text{CE}} \) is the Cross Entropy loss, \( \mathcal{L}_{\text{Cosine}} \) is the Cosine Similarity loss, and \( \mathcal{L}_{\text{Ortho}} \) is the Feature Orthogonality loss. The hyperparameters \( \alpha, \beta, \gamma, \delta \) control the contribution of each individual loss term defined as:
\[
\mathcal{L}_{\text{KL}} = D_{\text{KL}}(P \| Q) = \sum_{i} P(i) \log \frac{P(i)}{Q(i)}
\]
\[
\mathcal{L}_{\text{CE}} = -\sum_{i} y_{i} \log(\hat{y}_{i})
\]
\[
\mathcal{L}_{\text{Cosine}} = \frac{1}{N(N-1)} \sum_{i \neq j} \cos(\theta_{ij})
\]
\[
\mathcal{L}_{\text{Ortho}} = \frac{1}{N^2} \sum_{i, j} (G_{ij} - \delta_{ij})^2
\]
where \( D_{\text{KL}} \) represents the KL Divergence between the input distribution \( P \) and the output distribution \( Q \), \( y_{i} \) is the set encoded label, \( \hat{y}_{i} \) is the predicted label from the classifier, \( \cos(\theta_{ij}) \) represents the cosine similarity between features of generated images \( i \) and \( j \), \( G_{ij} \) is the element of the Gram matrix, and \( \delta_{ij} \) is the Kronecker delta function. \( N \) is the number of feature vectors in the batch.

Thus, the combined loss function ensures that the generator matches the input and output distributions using KL Divergence and also generates images with desired labels using Cross Entropy, while maintaining diversity in the generated images through Feature Orthogonality and Cosine Similarity.

\subsubsection{Cross Entropy}
The key goal of the inversion process is to generate images with the desired labels and the same can be easily achieved using cross entropy loss. In cases where the label information is encoded into the vectors without being explicitly revealed to the generator, the encoded labels can be used in the cross entropy loss function with the classifier outputs for the generated images in order to train the generator. In contrast to the label conditioning, vector conditioning complicate the training objectives to the extent that the generator does not immediately converge, instead the convergence occurs only when the generator figures out the encoded conditioning mechanism allowing for a better exploration of the input space of the classifier.

\subsubsection{KL Divergence}
KL Divergence is used to train the generator to learn the data distribution in the input space of the classifier for different conditioning vectors. During training, the KL Divergence loss function measures and minimise the difference between the output distribution of the generated images, as predicted by the classifier, and the conditioning distribution used to generate these images. This divergence metric is crucial for aligning the generated image distributions with the intended conditioning distribution.

\subsubsection{Cosine Similarity}
To enhance the diversity of the generated images, we use cosine similarity to assesses and minimises the angular distance between the features of a batch of generated images across the last fully connected layers, promoting variability in the generated images. The combination of cosine similarity with cross-entropy loss not only ensures that the generated images are classified correctly but also enforces diversity among the images produced for each label.

\subsubsection{Feature Orthogonality}
In addition to the cosine similarity loss, we incorporate feature orthogonality as a regularization term to further enhance the diversity of generated images by minimizing the deviation of the Gram matrix of the features from the identity matrix. By ensuring that the features of generated images are orthogonal, we promote the generation of distinct and non-redundant representations for each conditioning label. 

\begin{figure*}[h]
\centering
\includegraphics[width=1\textwidth]{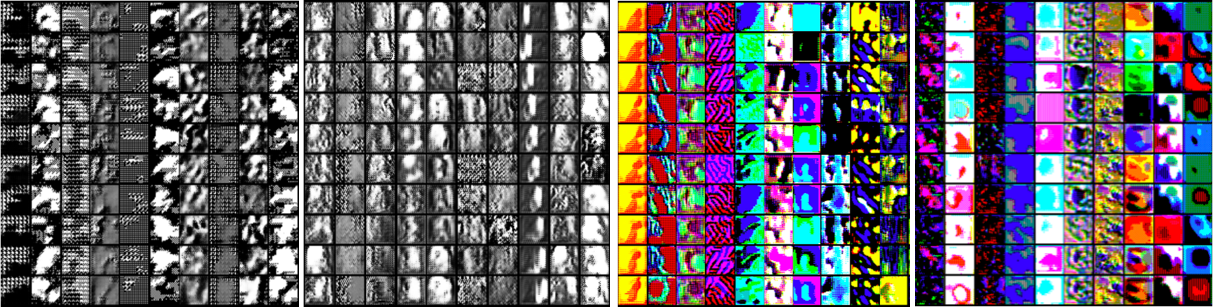}
\caption{Inverted Images for all 10 classes in MNIST, FashionMNIST, SVHN \& CIFAR-10 respectively.}
\label{2}
\end{figure*}

\section{Results}
\label{sec:res}
In this section, we present the experimental results obtained by applying our inversion technique on the MNIST \citep{deng2012mnist}, FashionMNIST \citep{xiao2017fashionmnistnovelimagedataset}, SVHN and CIFAR-10 \citep{cf} datasets by training a generator to produce images that, when passed through a classifier, elicit the desired labels. The classifier is initially normally trained on a dataset and then held in evaluation for the purpose of inversion and reconstruction. The images generated by the conditioned generator corresponding to the latent and the conditioning vectors are then passed through the classifier.

The classifier is a simple multi-layer convolutional neural network consisting of convolutional layers, dropout layers, batch normalization, and leaky-relu activation followed by fully connected layers and softmax for classification. While the generator is based on Vector-Matrix Conditioning in which the class labels are encoded into random softmaxed vectors concatenated with the latent vector followed by multiple layers of transposed convolutions, batch normalization \citep{pmlr-v37-ioffe15} and dropout layers \citep{JMLR:v15:srivastava14a} to encourage diversity in the generated images. Once the vectors are upsampled to \(NXN\) spatial dimensions for an N class classification task they are concatenated with a conditioning matrix for subsequent generation upto the required image size of 28X28 or 32X32.

The inverted images are visualized to assess the quality and diversity of the generated samples in Figure \ref{2} for all 10 classes of MNIST, FashionMNIST, SVHN and CIFAR-10 respectively. While each row corresponds to a different class each column corresponds to a different generator and as can be observed the images within each row represent the diversity of samples generated for that class. It is observed that high weightage to cosine similarity increases both the inter-class and the intra-class diversity in the generated samples of a single generator. These inverted samples that are confidently classified by the generator are unlike anything the model was trained on, and yet happen to be in the input space of different labels highlighting their unsuitability in safety-critical tasks.

\section{Applications}
\label{sec:apps}
In this section we study the applications of our proposed network inversion approach in interpretability, out-of-distribution (OOD) detection, and training data reconstruction.
\subsection{Interpretability}
\begin{figure*}[h]
\centering
\includegraphics[width=1\textwidth]{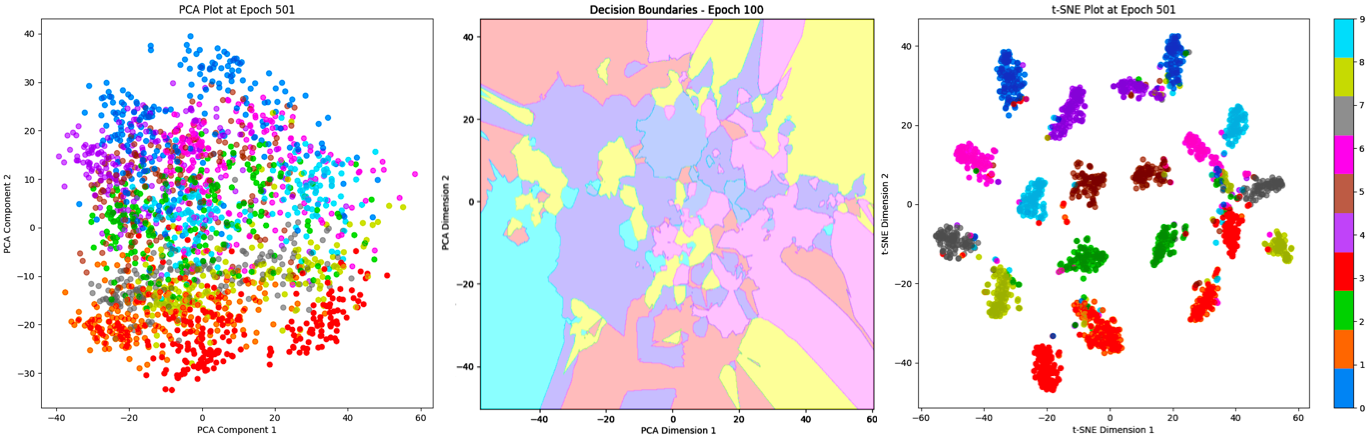}
\caption{PCA, Decision Boundaries and t-SNE plots of the features extracted from the generated images. Each color represents a different class. While the PCA plot illustrates the spread of a class in the feature space, t-SNE plot shows the two different clusters for each class.}
\label{3}
\end{figure*}

In interpretability, we analyze the features of the inverted samples generated after training the generator to an Inversion Accuracy of over 95\%. Inversion Accuracy refers to the percentage of images generated with desired labels same as the output labels from the classifier. Figure \ref{3} shows the PCA plots, decision boundaries and t-SNE plots respectively for the features in the penultimate layer of the classifier for the inverted samples.

To evaluate the diversity of features corresponding to each class, PCA (Principal Component Analysis) plots illustrate the distribution of class-specific features across the feature space. This spread aligns with our objective of generating diverse inputs that represent various characteristics within a particular class, ensuring a comprehensive understanding of each class’s feature space. We further visualize interpretable decision boundaries by mapping the PCA-transformed features onto a mesh grid and performing inference on this feature mesh grid allowing us to highlight the regions associated with different classes. This visualization reveals how features from inverted samples influence the network’s classification criteria, providing insight into the model’s internal decision-making structure.

Additionally, we employ t-SNE (t-distributed Stochastic Neighbor Embedding) plots to explore the distinct distributions from which the inverted images are generated. As can be observed from the t-SNE plots, for a specific value of the hyperparameters in our loss function, images for each class originate from two separate distributions. We also observe that increasing the weightage of the cosine similarity term in the loss function further enhances the diversity among inverted samples, leading to the emergence of multiple clusters within each class.

Lastly, we also use Sparse Autoencoders (SAEs) \citep{makhzani2014ksparseautoencoders}, \citep{7962256} on the features in the penultimate layer of the classifier for the inverted samples to identify interpretable feature patterns. While SAEs have traditionally been applied to analyze features of the actual training data, our application of SAEs on a diverse set of inverted samples reveals distinct feature activations for different data distributions generated by various generators within the same class. Notably, the set of features activated for the training data differs from those activated for random inverted samples even within the same class, suggesting that this approach could potentially be used in anomaly detection.

\subsection{Out-of-Distribution Detection}
Classifiers excel at distinguishing between different classes with high accuracy, but they often fail when confronted with out-of-distribution (OOD) samples—inputs that deviate from the training distribution. These counterexamples are frequently misclassified with high confidence, exposing a critical weakness in the model's generalization capabilities. Therefore, it is crucial to make classifiers more robust to counterexamples and enable them to flag OOD samples into a separate class.

To address this issue, we leverage network inversion to generate OOD samples from the classifier’s input space for each label. These inverted samples, which represent data outside the classifier’s learned distribution, are then assigned to a designated "garbage" class. This process allows the classifier to explicitly recognize and isolate spurious data points, improving its robustness and reliability.

To implement this approach, we begin by training the classifier with an extra "garbage" class that initially contains samples generated from random Gaussian noise. This step provides a baseline for the classifier to identify nonspecific, out-of-distribution patterns. After each training epoch, inverted samples generated for each class are added to this "garbage" class, further augmenting the dataset with a wider variety of OOD examples. This iterative retraining process helps the classifier better recognize and distinguish the boundary between legitimate (in-distribution) and spurious (OOD) data, improving its robustness in practical applications.

To manage the data imbalance introduced by the addition of the garbage class, we employ a weighted cross-entropy loss function. This function assigns varying weights to each class, which are dynamically adjusted after each epoch to reflect the evolving class distribution. This weighted approach ensures that the classifier continues to learn effectively from the augmented dataset while addressing any imbalance caused by the inclusion of inverted samples. By reinforcing the classifier’s ability to generalize and detect OOD samples, this method mitigates the influence of class imbalances, thereby enhancing overall robustness and accuracy in handling OOD detection tasks.

Out-of-distribution (OOD) detection experiments were conducted on models trained on MNIST and tested for OOD detection on FMNIST, and vice versa. Similarly, models trained on SVHN were tested on CIFAR10, and vice versa. These experiments evaluate the effectiveness of the proposed approach in identifying OOD samples and assigning them to the "garbage" class. In these experiments, we observe that while the majority of OOD samples are correctly assigned to the garbage class, a small percentage of the samples can still be misclassified into in-distribution classes. However, a significant finding is that the least confidently classified in-distribution sample is still more confidently classified compared to the most confidently misclassified out-of-distribution sample, suggesting the existence of a clear threshold. This clear distinction in confidence levels demonstrates the robustness of the proposed approach in separating in-distribution and OOD data, highlighting its potential for reliable OOD detection in practical scenarios.
\begin{figure*}[t]
\centering
\includegraphics[width=0.9\textwidth]{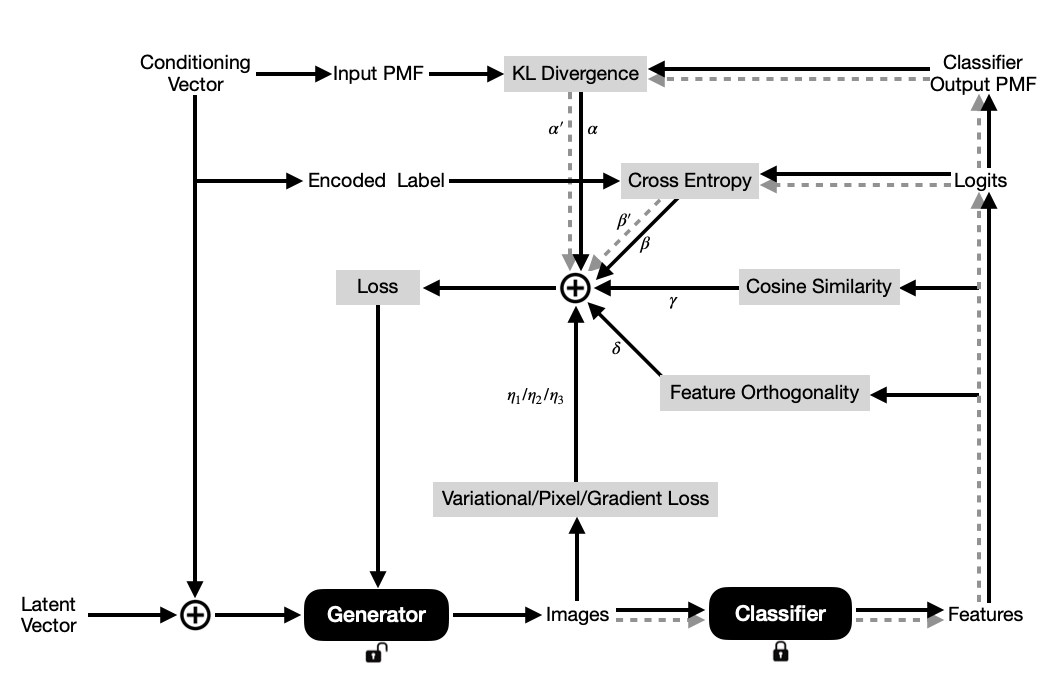} 
\caption{Schematic Approach to Training-Like Data Reconstruction using Network Inversion}
\label{4}
\end{figure*}

\subsection{Training-Like Data Reconstruction}

\begin{figure*}[h]
\centering
\includegraphics[width=1\textwidth]{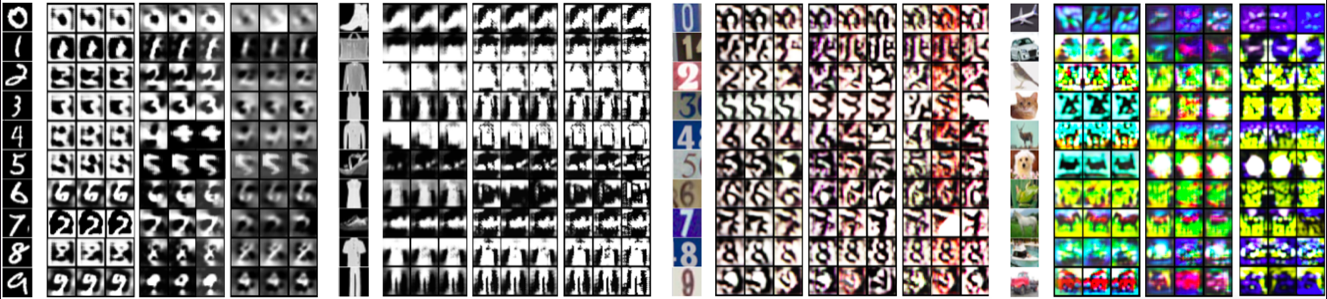}
\caption{Reconstructed Images for all 10 classes in MNIST, FashionMNIST, SVHN and CIFAR10 respectively .}
\label{5}
\end{figure*}

We also demonstrate the utility of our proposed network inversion technique for reconstructing training-like data entirely from the weights of the single classifier without any insight of the training process. The paper \citep{haim2022reconstructingtrainingdatatrained} introduces a novel reconstruction method based on the implicit bias of gradient-based training methods to reconstruct training samples specifically focusing on binary MLP classifiers. Later \citep{buzaglo2023reconstructingtrainingdatamulticlass} extend on these results by perfroming training data reconstruction in a multi-class setting on models trained on even larger number of samples. The paper \citep{9833677} addresses the issue of whether an informed adversary, who has knowledge of all training data points except one, can successfully reconstruct the missing data point given access to the trained machine learning model. Subsequenlty \citep{pmlr-v206-wang23g} investigates how model gradients can leak sensitive information about training data, posing serious privacy concerns.

While in \citep{haim2022reconstructingtrainingdatatrained} training data is reconstructed for binary multi-layer perceptron classifiers trained using binary cross entropy, we steer network inversion to reconstruct data similar to the training set for Convolutional Neural Nets.
In restricted settings, over-parameterized models can easily memorize portions of the training data, leading to successful reconstructions. For under-parameterized models, where there is no possibility of memorization and the models generalize well, reconstructions are typically more difficult. Also in fully connected layers, each input feature is assigned dedicated weights, which may make reconstruction easier as the model captures more direct associations between inputs and outputs. While as in convolutional layers, due to the weight-sharing mechanism, where the same set of weights is applied across different parts of the input, the reconstruction becomes more challenging.

Network Inversion can be used for training data reconstruction as shown schematically in Figure \ref{4} by exploiting key properties of the training data in relation to the classifier that guide the generator towards producing training-like data including model confidence, robustness to perturbations, and gradient behavior along with some prior knowledge about the training data.

In order to take model confidence into account, we use hot conditioning vectors in reconstruction instead of soft conditioning vectors used in inversion encouraging the generation of samples that elicit high-confidence predictions from the model. Since the classifier is expected to handle perturbations around the training data effectively, the perturbed images should retain the same labels and also be confidently classified. To achieve this, we introduce an \(L_\infty\) perturbation to the generated images and pass both the original and perturbed images represented by dashed lines, through the classifier and use them in the loss evaluation. We also introduce a gradient minimization loss to penalise the large gradients of the classifier’s output with respect to its weights when processing the generated images ensuring that the generator produces samples that have small gradient norm, a property expected of the training samples.
Furthermore, we incorporate prior knowledge through pixel constraint and variational losses to ensure that the generated images have valid pixel values and are noise-free ensuring visually realistic and smooth reconstructions.

Hence the previously defined inversion loss  \(\mathcal{L}_{\text{Inv}}\) is augmented to include the above aspects into a combined reconstruction loss \(\mathcal{L}_{\text{Recon}}\) defined as:
\begin{align*}
\mathcal{L}_{\text{Recon}} = & \; \alpha \cdot \mathcal{L}_{\text{KL}} 
+ \alpha' \cdot \mathcal{L}_{\text{KL}}^{\text{pert}}
+ \beta \cdot \mathcal{L}_{\text{CE}} 
+ \beta' \cdot \mathcal{L}_{\text{CE}}^{\text{pert}} \\
& + \gamma \cdot \mathcal{L}_{\text{Cosine}} 
+ \delta \cdot \mathcal{L}_{\text{Ortho}} \\
& + \eta_1 \cdot \mathcal{L}_{\text{Var}} 
+ \eta_2 \cdot \mathcal{L}_{\text{Pix}} 
+ \eta_3 \cdot \mathcal{L}_{\text{Grad}}
\end{align*}

where \(\mathcal{L}_{\text{KL}}^{\text{pert}}\) and \(\mathcal{L}_{\text{CE}}^{\text{pert}}\) represent the KL divergence and cross-entropy losses applied on perturbed images, weighted by \( \alpha'\) and  \(\beta' \)respectively while \(\mathcal{L}_{\text{Var}}\), \(\mathcal{L}_{\text{Pix}}\) and \(\mathcal{L}_{\text{Grad}}\) represent the variational loss, Pixel Loss and penalty on gradient norm each weighted by \( \eta_1\), \( \eta_2\), and \(\eta_3\) respectively and defined for an Image \(I\) as:
\begin{align*}
\mathcal{L}_{\text{Var}} = \frac{1}{N} \sum_{i=1}^{N} \Bigg( \sum_{h,w} & \Big( ( I_{i, h+1, w} - I_{i, h, w} )^2 \\
& + ( I_{i, h, w+1} - I_{i, h, w} )^2 \Big) \Bigg)
\end{align*}
\[
\mathcal{L}_{\text{Grad}} = \left\| \nabla_{\theta} L(f_{\theta}(I), y) \right\|
\]

\[
\mathcal{L}_{\text{Pix}} = \sum \max(0, -I) + \sum \max(0, I - 1)
\]
The reconstruction experiments were carried out on models trained on datasets of varying size and as a general trend the quality of the reconstructed samples degrades with increasing number of the training samples. In case of MNIST and FashionMNIST reconstructions were performed for models trained on datasets of size 1000, 10000 and 60000, while as for SVHN and CIFAR-10, on datasets of size 1000, 5000, and 10000.
The reconstruction results using three different generators on each of the three dataset sizes on all four datasets are shown in Figure \ref{5} along with a column of representative training data. In case of SVHN we held out a cleaner version of the dataset in which every image includes a single digit. While as in case of CIFAR-10 given the low resolution of the images the reconstructions in some cases are not perfect although they capture the semantic structure behind the images in the class very well.

\section{Conclusions}

This paper introduced a novel approach to network inversion, utilizing a conditioned generator to generate a diverse set of inputs with desired output labels. By shifting from simple label conditioning to vector encoding and incorporating heavy dropout during the generation process, our method complicates the conditioning mechanism, encouraging the generator to explore a more extensive range of the data distribution.

In interpretability, it provides insights into the internal decision-making patterns of neural networks by analyzing the features of inverted samples, visualizing decision boundaries, and identifying diverse feature representations through tools like PCA, t-SNE, and Sparse Autoencoders. For out-of-distribution (OOD) detection, network inversion is used to generate OOD samples that are assigned to a designated "garbage" class, improving the classifier's robustness by distinguishing in-distribution and OOD samples with a clear confidence threshold. Lastly, in training data reconstruction, the method reconstructs training-like data using the classifier’s weights, exploiting model confidence, gradient behavior, and prior knowledge to generate data semantically similar to the training data.

Future work will aim to quantify the aspects of the inversion technique and explore its potential in enhancing interpretability and robustness across various real-world tasks. 
{
    \small
    \bibliographystyle{ieeenat_fullname}
    \bibliography{main}
}


\end{document}